\documentclass[letterpaper, 10 pt, journal, twoside]{IEEEtran}

\usepackage{booktabs}
\usepackage{graphics} %
\usepackage{graphicx}
\usepackage{epsfig} %
\usepackage{mathptmx} %
\usepackage{times} %
\usepackage{amsmath} %
\usepackage{amssymb}  %
\usepackage{newtxmath}
\usepackage{cite}
\usepackage{url}

\usepackage{algorithm}
\usepackage{algorithmic}
\usepackage{mathtools}
\usepackage{wrapfig}

\usepackage{subcaption,siunitx,booktabs}

\usepackage[export]{adjustbox}

\usepackage{color}
\usepackage{xcolor}
\definecolor{citecolor}{HTML}{0071bc}
\usepackage[pagebackref=true,breaklinks=true,colorlinks,citecolor=citecolor,urlcolor=citecolor,bookmarks=false]{hyperref}

\newlength\savewidth

\DeclareMathOperator*{\argmin}{arg\,min}

\newcommand{\T}{\boldsymbol{T}}
\newcommand{\X}{\boldsymbol{X}}

\renewcommand{\H}{\boldsymbol{H}}

\renewcommand{\P}{\boldsymbol{P}}
\renewcommand{\b}{\boldsymbol{b}}
\newcommand{\x}{\boldsymbol{x}}
\newcommand{\y}{\boldsymbol{y}}
\newcommand{\z}{\boldsymbol{z}}
\newcommand{\ttheta}{\boldsymbol{\theta}}

\begin{document}

\title{Online Adaptation for Implicit Object Tracking and Shape Reconstruction in the Wild}

\author{Jianglong Ye$^{1}$, Yuntao Chen$^{2}$, Naiyan Wang$^{2}$, Xiaolong Wang$^{1}$%
\thanks{Manuscript received: February, 24, 2022; Revised May, 21, 2022; Accepted July, 02, 2022.}
\thanks{Jianglong Ye and Xiaolong Wang are with $^{1}$University of California San Diego. Yuntao Chen and Naiyan Wang are with $^{2}$TuSimple. Correspondence at {\tt\footnotesize xiw012@ucsd.edu}.}%
\thanks{This paper was recommended for publication by Editor Eric Marchand upon evaluation of the Associate Editor and Reviewers' comments.}
\thanks{Digital Object Identifier (DOI): see top of this page.}
}

\markboth{IEEE Robotics and Automation Letters. Preprint Version. Accepted July, 2022}
{Ye \MakeLowercase{\textit{et al.}}: Online Adaptation for Implicit Object Tracking and Shape Reconstruction in the Wild}

\maketitle

\begin{abstract}
  Tracking and reconstructing 3D objects from cluttered scenes are the key components for computer vision, robotics and autonomous driving systems. While recent progress in implicit function has shown encouraging results on high-quality 3D shape reconstruction, it is still very challenging to generalize to cluttered and partially observable LiDAR data. In this paper, we propose to leverage the continuity in video data. We introduce a novel and unified framework which utilizes a neural implicit function to simultaneously track and reconstruct 3D objects in the wild. Our approach  adapts the DeepSDF model (i.e., an instantiation of the implicit function) in the video online, iteratively improving the shape reconstruction while in return improving the tracking, and vice versa. We experiment with both Waymo and KITTI datasets and show significant improvements over state-of-the-art methods for both tracking and shape reconstruction tasks. Our project page is at \url{https://jianglongye.com/implicit-tracking}.
\end{abstract}


\IEEEpeerreviewmaketitle

\section{Introduction}
\IEEEPARstart{R}{ecent} development on implicit function has shown a tremendous success on high-quality 3D shape reconstruction~\cite{chen2019learning, mescheder2019occupancy, park2019deepsdf, chabra2020deep, jiang2020local}. For example, one representative work DeepSDF~\cite{park2019deepsdf} uses a deep auto-decoder that takes a shape code and a coordinate as inputs to predict the signed distance to the shape surface. However, it usually relies on pre-training on dense synthetic data  (e.g., ShapeNet~\cite{chang2015shapenet}, ScanNet~\cite{dai2017scannet}) to obtain category level prior, and then applying on similar distribution of data for deployment. However, how to collect such large-scale high-quality dense data becomes a burden in applying DeepSDF.

With the development of 3D sensing technologies, 3D point cloud video sequences become more commonly available. However, in such data, the object of interest in each frame is much sparser than that in the synthetic dataset due to viewing angle or distance. DeepSDF could hardly be applied directly on each single frame, since even with the learned prior, it either reconstructs close to mean shape or overfits to the noise given partial observations and the artifacts around. Fortunately, multi-frame data could make up this intrinsic defects, but calls for a robust algorithm that could jointly discover the poses of objects in each frame and utilize them for better shape reconstruction.

\begin{figure}
  \centering
  \includegraphics[width=\columnwidth]{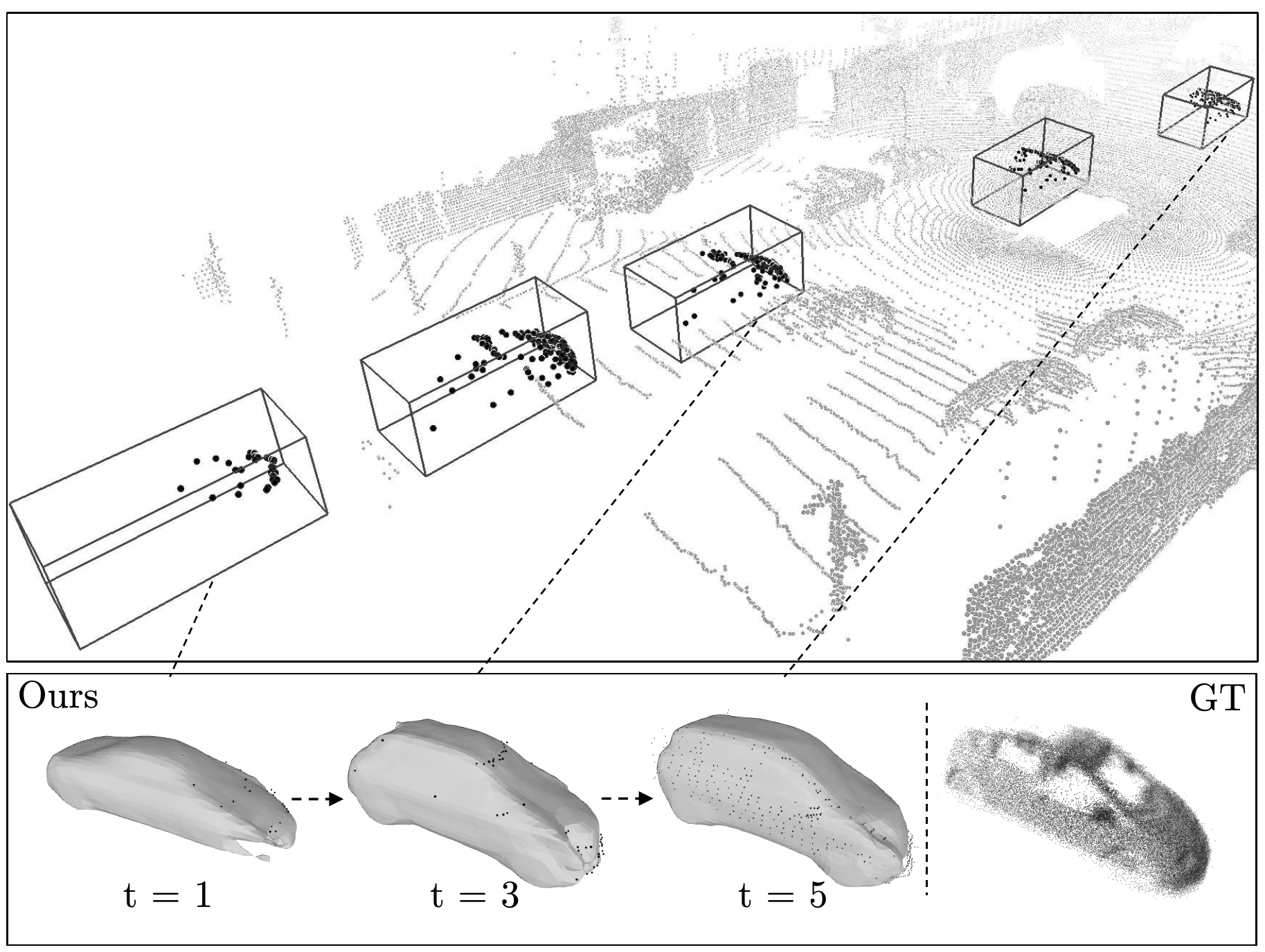}
  \vspace{-0.15in}
  \caption{\textbf{3D object tracking and shape reconstruction.} We propose a novel framework to utilize DeepSDF to perform tracking and reconstruction simultaneously. We visualize five time steps of tracking and three reconstructed shapes. We obtain better and better shape close to the ground-truth point clouds over time, which in return helps the tracking. }\label{fig: teaser}
  \vspace{-0.15in}
\end{figure}

In this paper, instead of taking object tracking and shape reconstruction as two separate tasks, we propose to solve them collaboratively using one single model. Specifically, we utilize the tracked poses of objects to update the DeepSDF codes, while the DeepSDF shape helps tracking in challenging cases such as sparse or occluded observations. Our framework iterates between object tracking and online adaptation along the video to improve performance for both tasks. As shown in Figure~\ref{fig: teaser}, the reconstructed shape improves progressively as more frames are tracked. Note that we apply DeepSDF as the baseline in our framework, because of its simplicity and the optimization nature of inference with the auto-decoder model. Our framework can be also extended with other deep implicit functions.

To track with DeepSDF, we first optimize the shape code via back-propagation using the localized partial point clouds given in the initial time step. While the shape code itself might not give perfect reconstruction, it offers a shape prior to construct a complete signed distance field of the object as a template. We then perform object tracking via optimizing a differentiable template matching process. Specifically, given the current model and point clouds in a new time step, we apply a 3D transformation on the partial point clouds (3D translation and rotation) and feed them to the DeepSDF model to compute their signed distances. If the 3D transformation is correct, the distance should be close to zero. We then optimize the absolute distance via back-propagation. Since the 3D transformation operation is differentiable, we can back-propagate the gradients through the point clouds to adjust the transformation. Once the object point clouds are localized with the correct pose, we can then use the point clouds to optimize the shape code.

We conduct our experiments with LiDAR video data recorded in the wild. Our DeepSDF model is first trained in the ShapeNet~\cite{chang2015shapenet} dataset to obtain the shape prior. We then perform online adaptation with the model on both Waymo~\cite{sun2020scalability} and KITTI~\cite{geiger2012we} datasets. We demonstrate that our method not only achieves state-of-the-art performance on 3D object tracking in both datasets, but also improves shape reconstruction at the same time.

We highlight our main contributions as follows:
\begin{itemize}
  \item We propose a novel framework that adapts online for simultaneous tracking and shape reconstruction and boosts the performances of both.
  \item We introduce the learning-based implicit function with its shape prior to this joint task and prove its effectiveness.
  \item Our method achieves the state-of-the-art performance in 3D single object tracking, while also improves the performance of shape reconstruction.
\end{itemize}

\section{Related Work}

\noindent
\textbf{Implicit neural representations.} In the past few years, a lot of advances have been made in 3D representation learning with meshes~\cite{bagautdinov2018modeling, wang2018pixel2mesh, defferrard2016convolutional}, point clouds~\cite{qi2017pointnet, yang2017foldingnet, yuan2018pcn} and voxel grids~\cite{wu20153d, choy20163d,girdhar2016learning}. Recently, the learning-based implicit functions have shown promising results on high-quality 3D shape reconstruction~\cite{park2019deepsdf, mescheder2019occupancy, chen2019learning, gropp2020implicit, xu2019disn, saito2019pifu, saito2020pifuhd, DBLP:conf/cvpr/GenovaCSSF20, duggal2022mending}.
For example, Mescheder et al.~\cite{mescheder2019occupancy} proposed the Occupancy network which learns to predict whether a 3D point is inside or outside the object given the 3D point cloud or image inputs. Saito et al.~\cite{saito2019pifu} further extended this work with local feature encoding beyond global shape features. Both of these approaches adopt an encoder-decoder architecture that reconstructs the shape in a feed-forward pass. In parallel, Park et al.~\cite{park2019deepsdf} introduced an auto-decoder DeepSDF for learning the signed distance to the shape surfaces. Instead of one forward pass, DeepSDF offers a shape code to optimize during inference and achieve better generalization on unseen instances. Based on this observation, Duggal et al.\cite{duggal2022mending} introduced good practices on generalizing DeepSDF to point cloud data in the wild.

\noindent
\textbf{3D object tracking.} There are two paradigms for 3D object tracking. The first paradigm adopts a tracking-by-detection strategy~\cite{weng20203d, simon2019complexer, luo2018fast, shenoi2020jrmot, baser2019fantrack} which utilizes an off-the-shelf detector\cite{ku2018joint, qi2018frustum} to localize the objects in each time step, and then associates the objects across frames for tracking. However, the performance of these methods high relies on the detection algorithm, and the temporal continuity in video data is not well explored. The second paradigm leverages the tracking results in previous frames and their similarities to the current frame for tracking~\cite{giancola2019leveraging, hu2019joint, zarzar2019efficient, qi2020p2b, fang20203d}. For example, Giancola et al.~\cite{giancola2019leveraging} first proposed to apply a Siamese network in 3D for matching and exploit a shape completion network for regularization.
While these approaches demonstrate their effectiveness, they still require  annotations in the video sequences to train the model. On the other hand, our method is annotation free except for the first frame. It directly adapts a model trained on synthetic data online for tracking. In this sense, our method is closely related to the setting in~\cite{ushani2015continuous, held2016robust, gross2019alignnet, pang2021model} where the pose is estimated by point cloud registration, and the shape could be obtained by aggregating point clouds from multiple frames. Instead of using aggregated point clouds as the shape template, our method utilizes an implicit neural representation that incorporates the shape prior from the whole category, allowing for more robust tracking.

\noindent
\textbf{Joint pose estimation and shape reconstruction.}  Our work is closely related to recent advances on joint estimation of 3D object pose and shape~\cite{zeeshan2014cars, engelmann2016joint, chabot2017deep, ke2020gsnet, ku2019monocular, najibi2020dops, zakharov2020autolabeling, zhang2021holistic, coenen2021pose}. For example, Zia et al.~\cite{zeeshan2014cars} and Coenen et al.~\cite{coenen2021pose} estimated the 3D wireframe models of multiple objects within a scene, which leads to higher performance in both object localization and viewpoint estimation. Ku et al.~\cite{ku2019monocular} predicted bounding boxes and instance point clouds from images simultaneously. Najibi et al.~\cite{najibi2020dops} and Zhang et al.~\cite{zhang2021holistic} employed implicit shape representation and proposed learning-based methods for joint 3D detection and shape prediction.  However, they have only focused on the estimation on a single frame. Going beyond a single image, Feng et al.~\cite{feng2019localization} utilized semantic mesh keypoints and segmentation masks to recover poses and shapes of cars in videos. Engelmann et al.~\cite{engelmann2017samp} adopted PCA on SDF volume to represent object shape and optimized poses and shape sequentially from video frames. While the general idea is relevant to our work, our method focuses on utilizing a more powerful representation with deep implicit functions.
To our knowledge, our method is the first work using the deep implicit function to simultaneously track 3D object and recover shape on in-the-wild videos.

\section{Method}

\begin{figure*}
  \centering
  \includegraphics[width=0.8\linewidth]{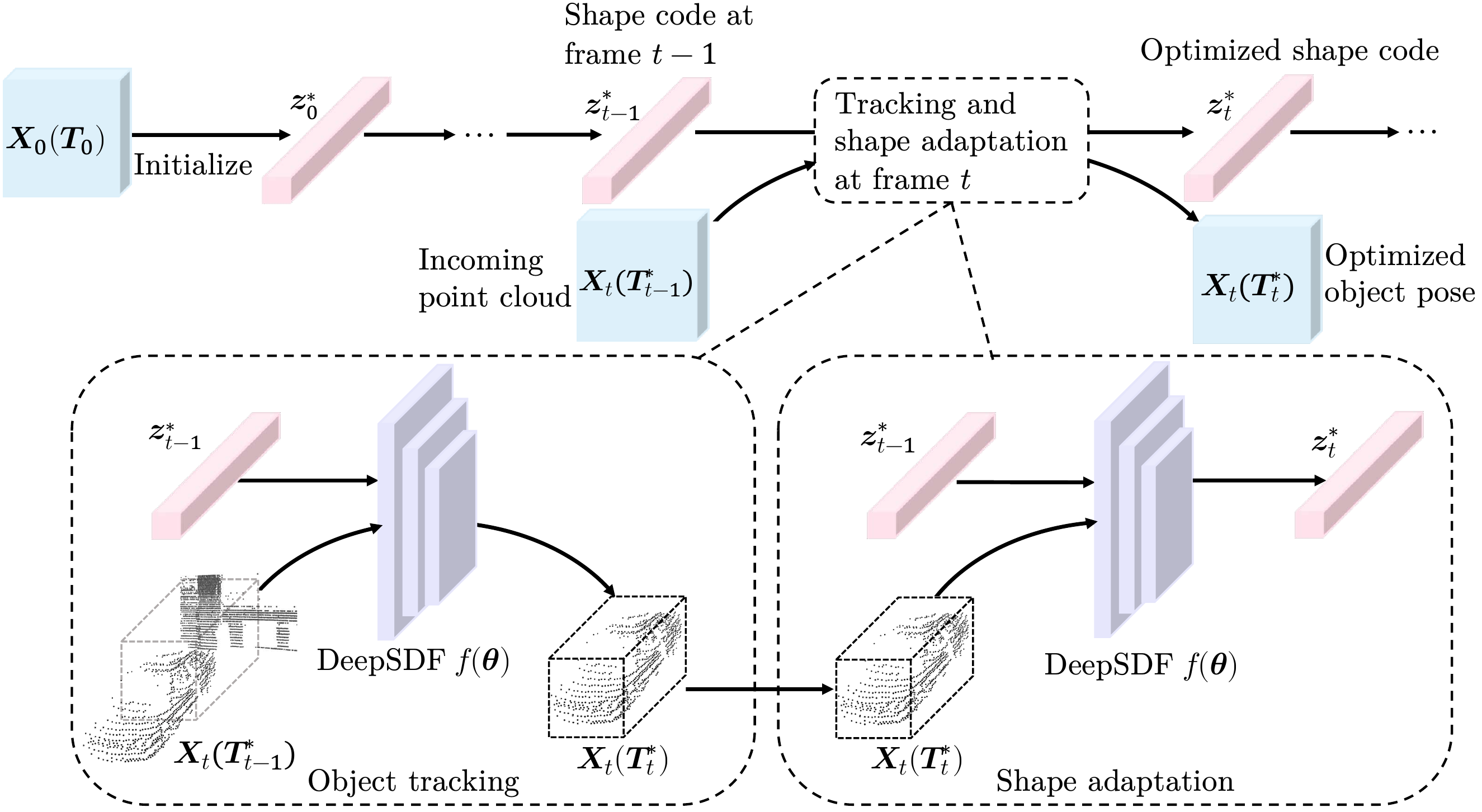}
  \vspace{-0.05in}
  \caption{\textbf{Overview of our method.} After initialization of the shape code, tracking and shape adaptation are performed iteratively. At a specific frame, the incoming object point cloud is first aligned to the previous shape, and then the shape is adapted to the aligned point cloud. Both procedures are based on DeepSDF.}\label{fig: method}
  \vspace{-0.1in}
\end{figure*}

\subsection{Pipeline}
Our method consists of three components: shape initialization, pose estimation and shape adaptation. %
At the beginning of single object tracking, a 3D bounding box of the object with pose $\T\in SE(3)$ and size $\b = \{h, w, l\} \in \mathbb{R}^3$ (height, width and length) are provided as the tracking template.
Note that, similar to mainstream 3D detection and tracking benchmarks, object pose here only consists of yaw and translation.
We use $\X(\T) = \{\x | \T \x \leq \b, \x\in \P\}$ to denote LiDAR points $\x$ in frame $\P$ that are inside the 3D bounding box.
We omit $\b$ for simplicity as the dimension of object is known and fixed.
As shown in Figure~\ref{fig: method}, we first initialize the shape with aligned point cloud at first frame, then in each frame we iterate between aligning the incoming object points $\X(\T)$ with reconstructed shape and adapting the shape from previous tracked point clouds $\{ \X_0(\T^*_0), \X_1(\T^*_1), \cdots, \X_t(\T^*_t)\}$ with optimal poses $\{ \T^*_0, \T^*_1, \cdots, \T^*_t \}$.

\subsection{Implicit Function for Shape Representation}
Vehicles in the wild come with a wide variation of shapes and partial observations as shown in Figure~\ref{fig: teaser}, which introduces challenges to the use of fixed shape models of car~\cite{engelmann2016joint, ke2020gsnet}.
Consequently, we choose DeepSDF~\cite{park2019deepsdf} as our shape representation for its expressiveness and flexibility. Using an auto-decoder to predict the signed distance to the surface, DeepSDF is able to represent shapes of arbitrary typologies with unlimited resolution. The continuous nature of DeepSDF allows our method to optimize object pose smoothly via back-propagation.

Signed distance function (SDF) is a widely used representation for 3D shape.
SDF maps the spatial coordinates of a point $\x$ to its signed distance $s$ to the closest surface.
\begin{equation}
  SDF(\x) = s,\quad\x\in\mathbb{R}^3, s\in\mathbb{R}.
\end{equation}

DeepSDF is a coordinate-based MLP $f$ parameterized by $\ttheta$ which approximates the SDF. It takes a 3D coordinate $\x$ and a learnable per-object shape code $\z$ as inputs. The DeepSDF function can be represented as,
\begin{equation}
  \label{deepsdf}
  f(\x, \z; \ttheta) = s,\quad \x\in\mathbb{R}^3, \z\in\mathbb{R}^d, s\in\mathbb{R},
\end{equation}
which encodes a one-to-one mapping between shape code $\z$ and 3D shape, and shape representation for different objects can be achieved by optimizing the shape code for each object, while leaving the parameters of MLP $\ttheta$ fixed.

In our method, we first pre-train the DeepSDF network using the \emph{car}   category of the ShapeNet Core~\cite{chang2015shapenet} dataset. Then, given the initial pose  $\T_0$, the corresponding optimal code (a.k.a shape representation) $\z^*_0$ could be obtained by
\begin{equation}
  \label{eqn: init}
  \z_0^* = \argmin_{\z} \sum_{\x\in\X_0(\T_0)} \mathcal{L}_\delta\left(f(\x, \z;\ttheta), 0\right) + \lambda \|\z\|_2^2,
\end{equation}
where $\mathcal{L}_\delta$ is the standard smooth $\ell_1$ loss, and the threshold $\delta$ is set as 0.05 in all experiments.
An $\ell_2$ regularizer is applied for the shape code $\z$ to prevent over-fitting.
Generally, points in a LiDAR scan come from the surface of objects, where the SDF values are 0.

\subsection{Joint Tracking and Shape Reconstruction in the Wild}

Different from standalone shape reconstruction where dense surface points of objects are already well-posed in the canonical view, our setting requires joint pose estimation and shape estimation, making it far more challenging.

Our joint tracking and shape reconstruction problem consists of three terms: The first is the SDF loss function, while the second is the regularization of the shape code to alleviate overfitting issue caused by partial view and noisy points. And the last one is the point cloud registration loss to capture fine-grained details. For frame $t$, we have:

\begin{equation}
    \min_{\T, \z} \sum_{\x\in\H_t\cup\X_t(\T)} \mathcal{L}_\delta \left(f(\x, \z;\ttheta), 0\right) + \gamma \mathcal{L}_c\left(\X_t(\T), \H_t\right)  + \lambda \|\z\|_2^2,
  \label{eqn: joint}
\end{equation}
where $\H_t$ stands for points from all previously tracked objects, which is defined as:
\begin{equation}
  \H_t =  \{ \X_0(\T_0^*) \cup \X_1(\T_1^*) \cup \cdots \cup \X_{t-1}(\T_{t-1}^*) \}.
\end{equation}

Here $\mathcal{L}_c(\X_t(\T), \H_t) = \sum_{\x\in\X_t(\T)} \min_{\y\in\H_t} \|\x - \y\|_2^2$ is the single-side Chamfer distance loss. The Chamfer distance loss between the object points in current frame and aggregated object points from previous time steps helps the pose estimation capture finer details of the object shape, as reconstructed shapes are tend to be over-smoothed.

Directly solving Eq.~\ref{eqn: joint} leads to sub-optimal results in practice as the pose $\T$ and the shape code $\z$ are two sets of variables with different scales and loss surfaces.
Instead, we take an iterative optimization approach as described in Algorithm~\ref{algo::track} and Figure~\ref{fig: method} to ease the difficulty of optimization. Specifically, our approach iterates between the \textbf{object tracking} task and \textbf{shape adaptation} task as follows.

\begin{figure}[t]
  \centering
  \begin{subfigure}{\columnwidth}
    \centering
    \includegraphics[width=0.9\columnwidth]{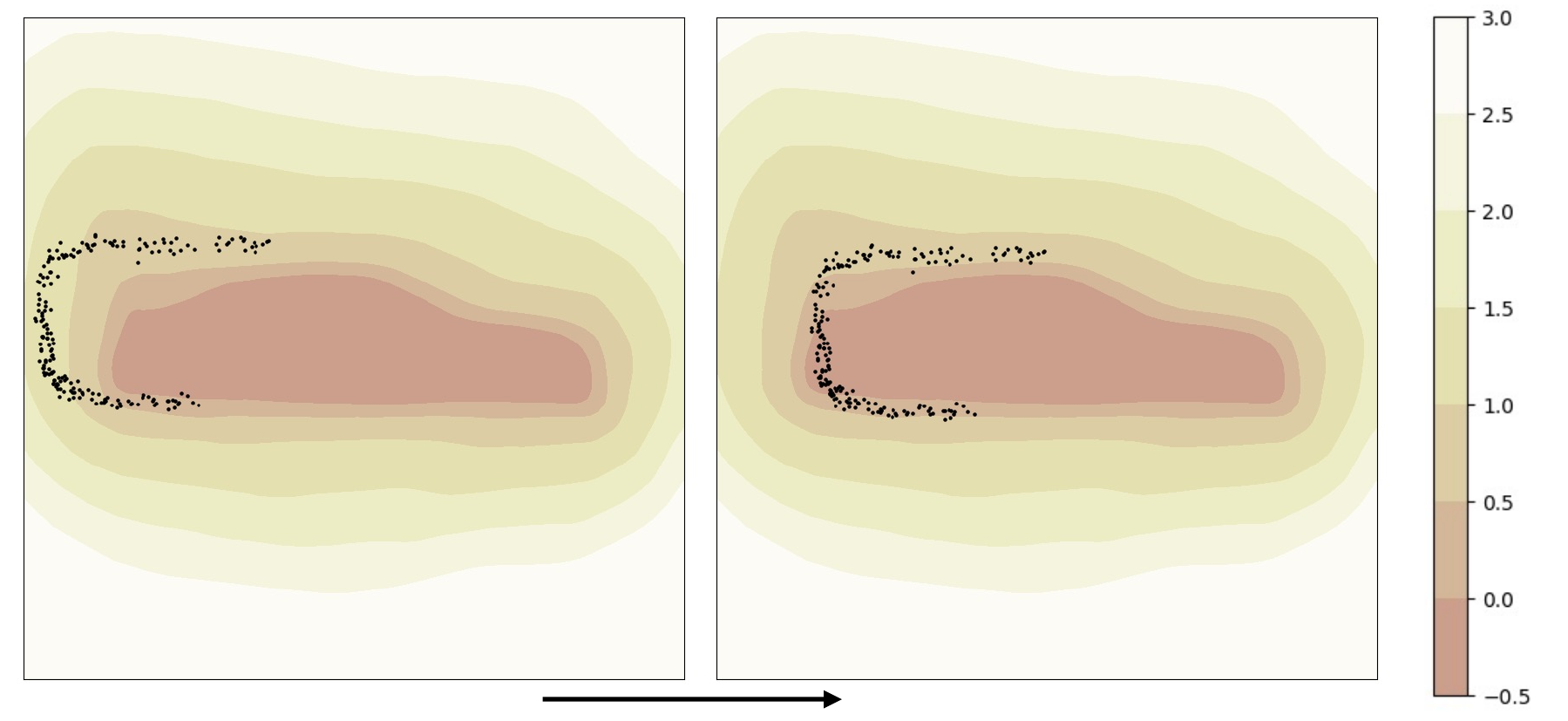}
    \vspace{-0.05in}
    \caption{Object tracking}
    \label{fig: field_tracking}
  \end{subfigure}
  \begin{subfigure}{\columnwidth}
    \centering
    \includegraphics[width=0.9\columnwidth]{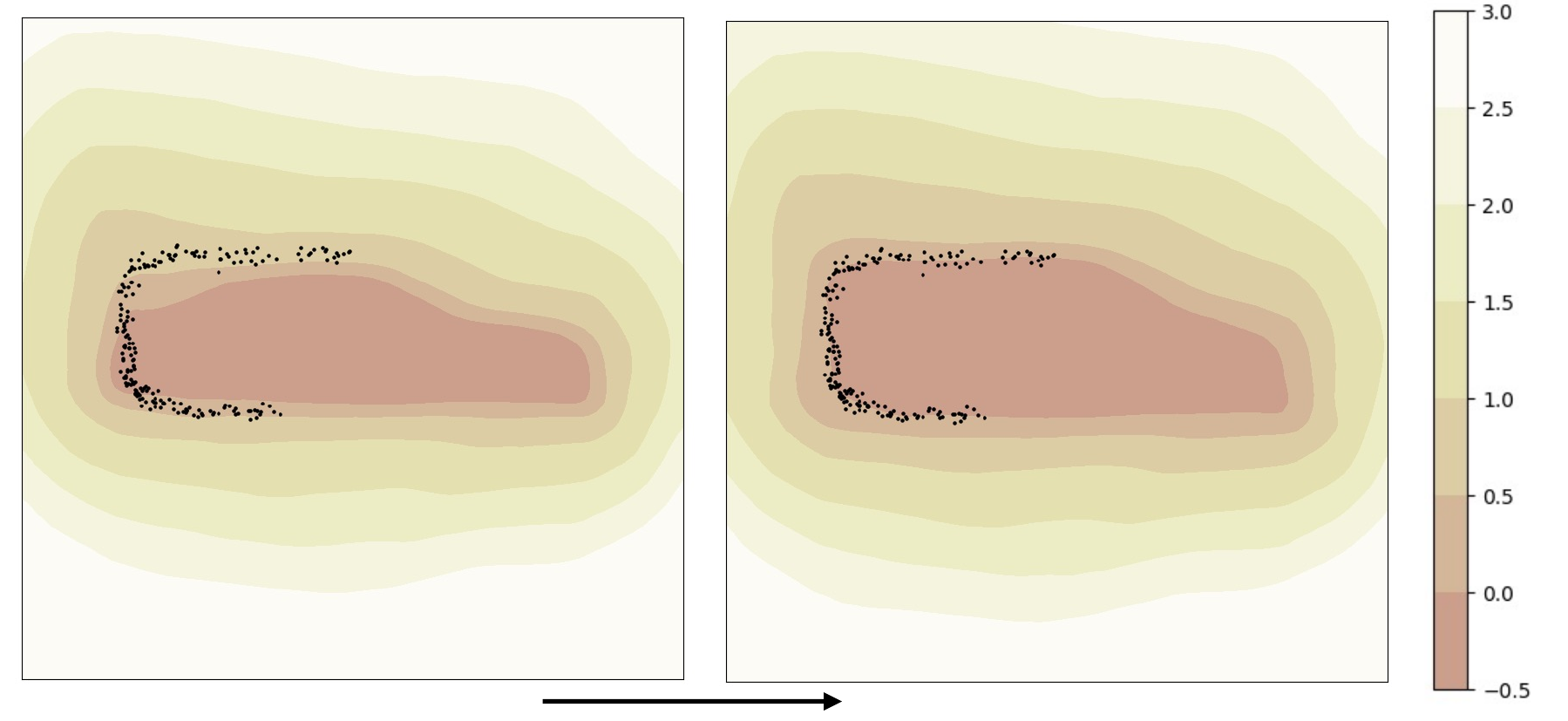}
    \vspace{-0.05in}
    \caption{Shape adaptation}
    \label{fig: field_shape}
  \end{subfigure}
  \vspace{-0.05in}
  \caption{\textbf{Iterative optimization.} We optimize pose by pushing the point cloud to the zero-level set of the SDF field in (a), and deform the SDF field to match the point cloud in (b) for better shape. We use different temperature of the color to represent the distances from the surface.}
  \label{fig: field}
  \vspace{-0.05in}
\end{figure}

\vspace{0.05in}
\noindent
\textbf{Object Tracking:}
To perform tracking, we fix the optimal shape code $\z^*_{t-1}$ obtained from the last frame, and optimize the object pose $\T_{t}$ as:
\begin{equation}
  \label{eqn: tracking}
  \T_t^* = \argmin_{\T} \sum_{\x\in\X_t(\T)} \mathcal{L}_\delta \left(f(\x, \z^*_{t-1};\ttheta), 0\right) + \mathcal{L}_c \left(\X_t(\T), \H_t\right),
\end{equation}

\noindent
which is actually Eq.~\ref{eqn: joint} with $\z$ fixed. We visualize the tracking process based on Eq.~\ref{eqn: tracking} in Figure~\ref{fig: field_tracking}. Initially, the point clouds are in the areas (the yellow zone) with high signed distance. By optimizing Eq.~\ref{eqn: tracking}, we push the point clouds towards the zero-level set (i.e., the surface of the shape template, boundary of the red zone). When optimizing the object pose, the loss weights of translation and rotation are equal.

\noindent
\textbf{Shape Adaptation:}
After aligning the point clouds, we adapt the shape utilizing observations to improve the shape quality. Specifically, we align the shape and the historical observations by minimizing the distances between them. The shape adaptation is achieved via optimizing the shape code,
\begin{equation}
  \label{eqn: adaptation}
  \z_t^* = \argmin_{\z} \sum_{\x\in\H_{t+1}} \mathcal{L}_\delta(f(\x, \z;\ttheta), 0) + \lambda \|\z\|_2^2,
\end{equation}
which is Eq.~\ref{eqn: joint} with $\T$ fixed.
We visualize the shape adaptation based on Eq.~\ref{eqn: adaptation} in Figure~\ref{fig: field_shape}. Specifically, given the tracked object location and pose, we deform the SDF field so that we push the boundary of the zero-level set (the boundary of the red zone) towards the point clouds. In return, this fine-tuned shape code could serve as a prior for tracking.

\noindent
\textbf{Detection Loss:} Since our method does not require training on a specific tracking dataset, it does not take advantage of human annotation on it. To leverage the information of the annotated bounding box, we can optionally incorporate an off-the-shelf detector. Specifically, we first select the closest detection result within a threshold of 3m as the corresponding predicted pose $\T_{det}$. Then we add a detection loss to minimize the difference between our estimated pose and the corresponding pose of detection. If there is no such detection box exists, then the detection loss is discarded. With the detection term, the Eq.~\ref{eqn: tracking} for object tracking  becomes:
\begin{equation}
  \label{eqn: detection}
  \begin{aligned}
    \T_t^* = \argmin_{\T} & \sum_{\x\in\X_t(\T)}  \mathcal{L}_\delta\left(f(\x, \z^*_{t-1};\ttheta), 0\right)  + \mathcal{L}_{\ell_1}\left(\T, \T_{det}\right)
    \\ & + \gamma \mathcal{L}_c\left(\X_t(\T), \H_t\right),
  \end{aligned}
\end{equation}
where $\mathcal{L}_{\ell_1}$ is the standard $\ell_1$ loss.

\begin{figure}[t]
\vspace{-0.15in}
\begin{algorithm}[H]
  \small
  \renewcommand{\algorithmicrequire}{\textbf{Input:}}
  \renewcommand{\algorithmicensure}{\textbf{Output:}}
  \caption{Pipeline}
  \label{algo::track}
  \begin{algorithmic}[1]
    \REQUIRE LiDAR scans $\P_{0:t}$, initial object pose $\T_0$, object size $\b$, DeepSDF $f$ with parameters $\ttheta$
    \ENSURE Subsequent object poses: $\T_{1:t}$, shape code $\z$
    \STATE Extract object points with initial pose and fixed object size: $\X_0(\T_0) \xleftarrow{} \{\x | \T_0 \x \leq \b, \x\in \P_0\}$. %
    \STATE Initialize shape code $\z_0$ by solving Eq.~\ref{eqn: init} with $\X_0(\T_0)$ and $f(\ttheta)$
    \STATE Frame number $t \xleftarrow{} 1$
    \WHILE{not the end of tracking}
    \STATE Extract object points with optimized pose at previous frame: $\X_t(\T_{t-1}^*) \xleftarrow{} \{\x | \T_{t-1}^* \x \leq \b, \x\in \P_t\}$
    \STATE Estimate object pose $\T_t^*$ by solving Eq.~\ref{eqn: tracking} with $\X_t(\T_{t-1}^*)$, $\z_{t-1}^*$ and $f(\ttheta)$.
    \STATE Update shape code $\z_t^*$ by solving Eq.~\ref{eqn: adaptation} with $\X_t(\T_{t}^*)$ and $f(\ttheta)$.
    \STATE $t \xleftarrow{} t + 1$
    \ENDWHILE
  \end{algorithmic}
\end{algorithm}
\vspace{-0.2in}
\end{figure}

\section{Experiments}

We conduct experiments on the Waymo open dataset~\cite{sun2020scalability} and the KITTI tracking dataset~\cite{geiger2012we} to evaluate both object tracking and shape reconstruction. We also perform thorough ablation studies to demonstrate how tracking and reconstruction can contribute to each other using online adaptation.

\begin{table*}[t]
  \centering

  \begin{tabular}{lcccccccc}
    \toprule
    & \multicolumn{2}{c}{All} & \multicolumn{2}{c}{ Easy } & \multicolumn{2}{c}{Medium} & \multicolumn{2}{c}{Hard} \\
    \texttt{Method} & Succ $\uparrow$ & Prec $\uparrow$ & Succ $\uparrow$ & Prec $\uparrow$ & Succ $\uparrow$ & Prec $\uparrow$ & Succ $\uparrow$ & Prec $\uparrow$ \\
    \midrule
    SOTracker~\cite{pang2021model} & 57.4 & 64.9 & 69.4 & \textbf{75.7} & 53.6 & 60.7 & 47.1 & 56.5 \\
    Ours & \textbf{62.3} & \textbf{65.7} & \textbf{71.5} & 74.1 & \textbf{58.8} & \textbf{61.8} & \textbf{54.9} & \textbf{59.9} \\
    \bottomrule
    \\[-0.05in]
    \toprule
    & \multicolumn{2}{c}{All} & \multicolumn{2}{c}{ Easy } & \multicolumn{2}{c}{Medium} & \multicolumn{2}{c}{Hard} \\
    \texttt{Method} & Acc $\uparrow$ & Rob $\uparrow$ & Acc $\uparrow$ & Rob $\uparrow$ & Acc $\uparrow$ & Rob $\uparrow$ & Acc $\uparrow$ & Rob $\uparrow$ \\
    \midrule
    SOTracker~\cite{pang2021model} & 57.1 & 50.0 & 69.2 & \textbf{64.5} & 53.2 & 45.1 & 46.6 & 37.8 \\
    Ours & \textbf{62.4} & \textbf{54.1} & \textbf{70.0} & 63.8 & \textbf{59.8} & \textbf{50.4} & \textbf{55.9} & \textbf{46.9} \\
    \bottomrule
    \\[-0.05in]
    \toprule
    & \multicolumn{2}{c}{All} & \multicolumn{2}{c}{ Easy } & \multicolumn{2}{c}{Medium} & \multicolumn{2}{c}{Hard} \\
    \texttt{Method} & ACD $\downarrow$ & Recall $\uparrow$ & ACD $\downarrow$ & Recall $\uparrow$ & ACD $\downarrow$ & Recall $\uparrow$ & ACD $\downarrow$ & Recall $\uparrow$ \\
    \midrule
    SOTracker~\cite{pang2021model} & 2.81 & 82.25 & 2.49 & 84.51 & 2.87 & 82.31 & 3.09 & 79.92 \\
    Ours & \textbf{2.50} & \textbf{84.71} & \textbf{2.30} & \textbf{86.40} & \textbf{2.76} & \textbf{83.78} & \textbf{2.44} & \textbf{83.98} \\
    \bottomrule
  \end{tabular}

  \caption{\textbf{Quantitative evaluation on Waymo.} From top to bottom, we present 1) comparison of tracking performance under Success and Precision metrics, 2) comparison of tracking performance under Accuracy and Robustness metrics, and 3) comparison of shape reconstruction performance under ACD and Recall metrics. $\uparrow$($\downarrow$) indicates performance is better with larger (smaller) values. Both SOTracker and our method do not employ the detection loss on Waymo.
  }
  \label{tab: comparison_waymo}
  \vspace{-0.1in}
\end{table*}

\subsection{Experimental Settings}
\label{sec: settings}

\noindent
\textbf{Pre-training.} For the pre-training of DeepSDF model, we use 2364 synthetic objects from the “car” category of the ShapeNet Core~\cite{chang2015shapenet} dataset. Since we aim to recover object shapes from partial observations, we simulate partial point cloud from LiDAR scan as the observation and use the complete SDF samples as the supervision of our DeepSDF model. For the observation, we randomly select 24 poses around each object which are 4-10 meters away from the object.  Based on selected poses, we sample partial point clouds on the mesh surface with the ray casting algorithm. For the supervision, we directly employ the processed SDF samples provided by DISN~\cite{xu2019disn}.

\noindent
\textbf{Datasets.}
For tracking on the Waymo dataset, we follow the protocol in LiDAR-SOT~\cite{pang2021model}, which utilizes 1121 tracklets that are longer than 100 frames and have more than 20 points in the first 10 frames. Besides, the dataset is spilt into easy, medium, and hard subsets based on the number of points in the first frame of each tracklet.
We also follow LiDAR-SOT~\cite{pang2021model} to remove the ground points to improve the performance. For tracking on the KITTI dataset, we follow the settings in SC3D\cite{giancola2019leveraging}. Since we do not need to train on KITTI, we only use scenes 17-18 from all 21 scenes for validation, 19-20 for testing.

\noindent
\textbf{Evaluation metrics.}  To evaluate the shape reconstruction, we aggregate point clouds from all frames of each tracklet using the annotated bounding boxes as pseudo-ground-truth shapes. Since the ground-truth aggregated points are not complete, we adopt Asymmetric Chamfer Distance (ACD) to measure the shape ﬁdelity. The ACD is defined as the mean squared distance from each ground-truth point to the nearest surface point on the reconstructed shape.
In addition, we calculate the recall at threshold $t$. The recall is defined as the fraction of ground-truth aggregated points within $t$ of a predicted point.
In the paper, we set threshold $t = 0.2m$.

For single object tracking, we adopt the Success and Precision metrics from  SC3D\cite{giancola2019leveraging}.
We add the Accuracy and Robustness metrics from LiDAR-SOT~\cite{pang2021model} for the Waymo dataset.

\noindent
\textbf{Implementation details.} We adapt DeepSDF as our shape model. Specifically, we reduce the MLP from 7 layers to 5 layers and remove the final \texttt{tanh} since our expected SDF is not truncated like DeepSDF. The dimension of the latent shape code is 512.
For the pre-training of DeepSDF, we use the Adam optimizer with an initial learning rate of $1 \times 10^{-4}$ with a batch size of 128.
During joint tracking and reconstruction, we use the SGD optimizer to optimize both object pose and shape code. The initial learning rates for pose estimation and shape code optimization are 0.1 and $1 \times 10^{-3}$, respectively. The numbers of iterations for pose estimation and shape code optimization are 300 and 20, respectively. In all experiments, the weight of regularization term $\lambda$ in  Eq.\ref{eqn: adaptation} is set to 10, and the weight of chamfer distance loss $\gamma$ in Eq.\ref{eqn: tracking} is set to 0.1.

Since the effect of online adaptation highly relies on the quality of the observations, it is necessary to filter out the poor quality observations. Thus we do not update the shape if the number of points in the frame is less than 10.

\begin{table}[t]
  \centering
  \begin{tabular}{lcccc}
    \toprule
    \texttt{Method} & Succ $\uparrow$ & Prec $\uparrow$ & ACD $\downarrow$ & Recall $\uparrow$ \\
    \midrule
    SC3D~\cite{giancola2019leveraging} & 41.3 & 57.9 & 1.68 & 88.56 \\
    P2B~\cite{qi2020p2b} & 56.2 & 72.8 & 1.46 & 91.08 \\
    3D-SiamRPN~\cite{Fang20213DSiamRPNAE} & 58.2 & 76.2 & / & / \\
    MLVSNet~\cite{wang2021mlvsnet} & 56.0 & 74.0 & 1.45 & 91.12 \\
    BAT~\cite{zheng2021box} & 60.5 & 77.7 & / & / \\
    SOTracker~\cite{pang2021model} & 43.9 & 59.4 & 1.64 & 89.01 \\
    Ours & \textbf{62.3} & \textbf{77.7} & \textbf{1.30} & \textbf{92.50} \\
    \hline
    VoxelRCNN + KF & 58.68 & 74.6 & 1.41 & 91.67 \\
    \bottomrule
  \end{tabular}
  \caption{\textbf{Quantitative evaluation on KITTI.} We outperform previous methods both in tracking (Success and Precision) and shape reconstruction (ACD and Recall).  Both SOTracker and our method employ the detection loss on the KITTI dataset. }
  \label{tab: comparison_kitti}
  \vspace{-0.1in}
\end{table}

\begin{figure*}[t]
  \centering
  \begin{subfigure}{\textwidth}
    \centering
    \includegraphics[width=0.95\textwidth]{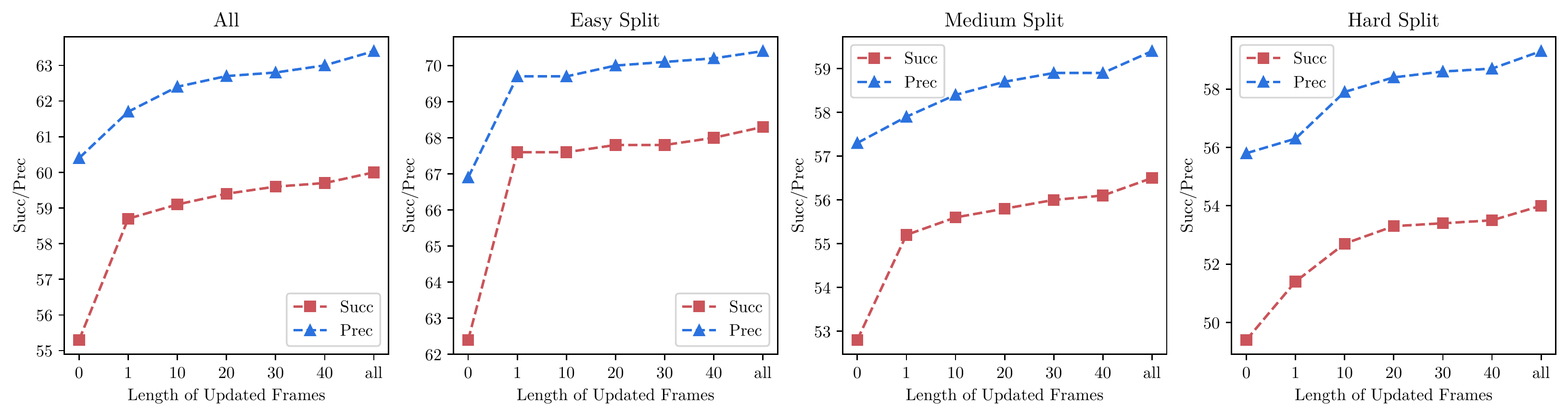}
    \vspace{-0.1in}
    \caption{Tracking performance}
    \label{fig: adaptation_tracking}
  \end{subfigure}
  \begin{subfigure}{\textwidth}
    \centering
    \includegraphics[width=0.95\textwidth]{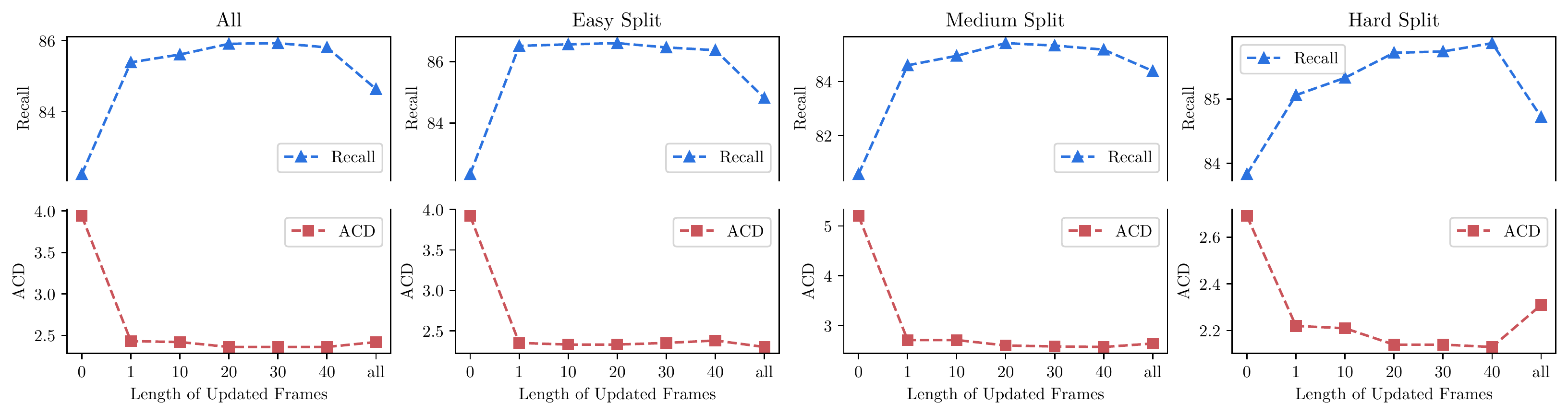}
    \vspace{-0.1in}
    \caption{Shape performance}
    \label{fig: adaptation_shape}
  \end{subfigure}
  \vspace{-0.05in}
  \caption{\textbf{Effectiveness of online adaptation mechanisms.} Due to the online adaptation mechanism, both tracking(a) and reconstruction(b) performance could be boosted. }
  \vspace{-0.15in}
  \label{fig: adaptation}
\end{figure*}

\begin{figure}[t]
  \centering
  \begin{subfigure}{0.7\columnwidth}
    \includegraphics[height=4.3cm, right]{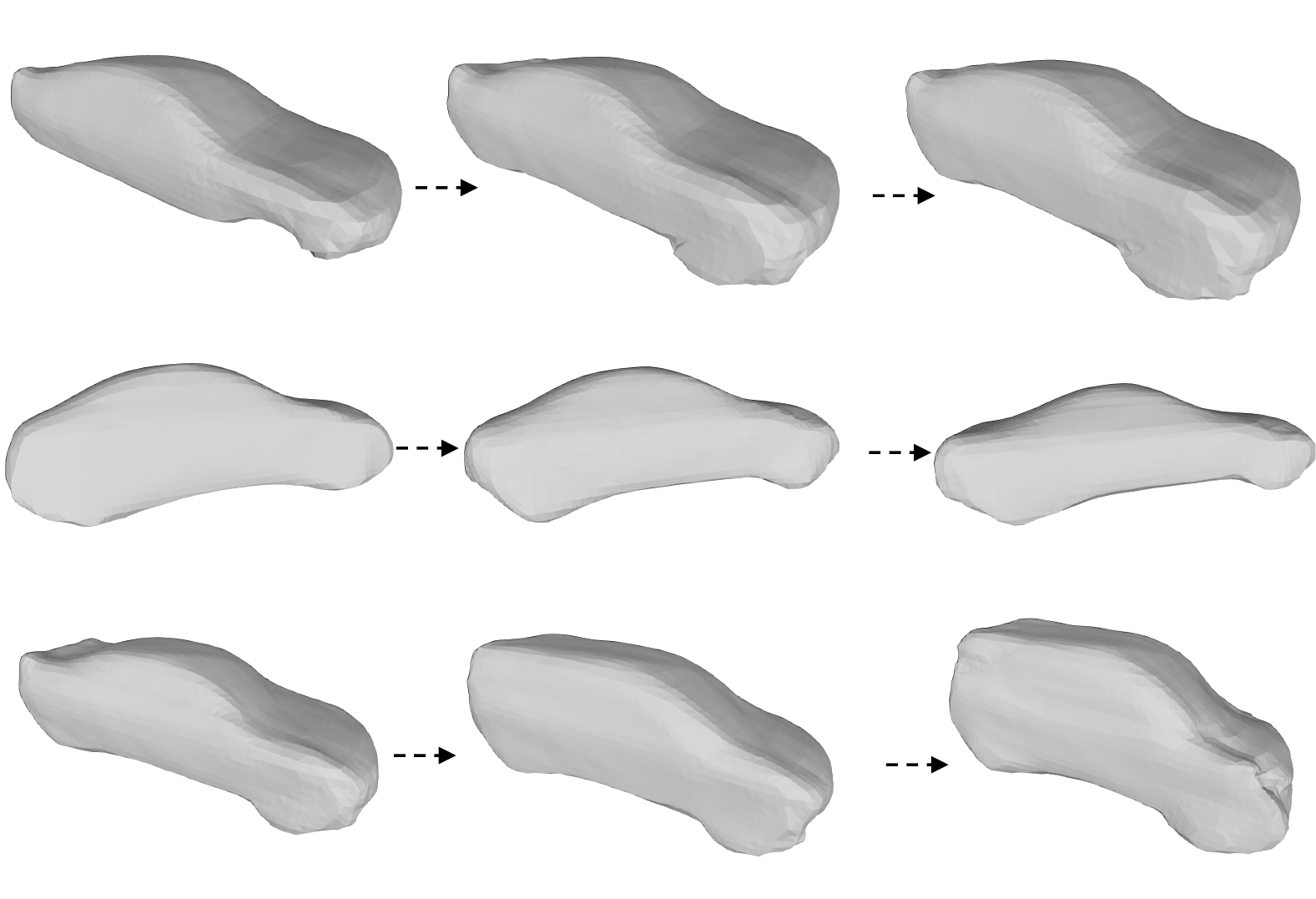}
    \caption{Shape evolution}
  \end{subfigure}
  \begin{subfigure}{0.2\columnwidth}
    \includegraphics[height=4.3cm, left]{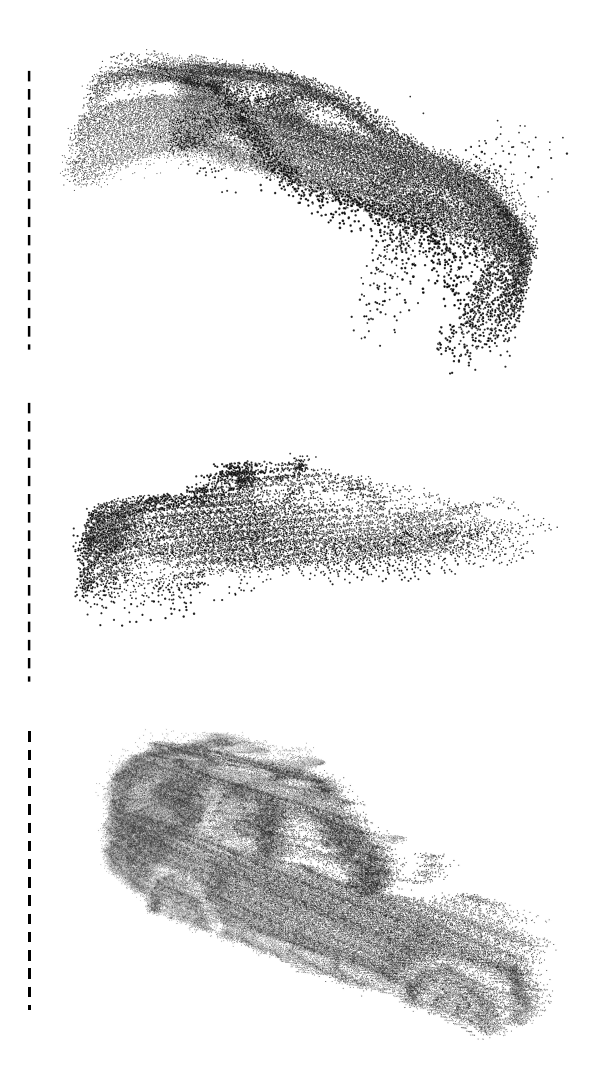}
    \caption{GT}
  \end{subfigure}
  \vspace{-0.05in}
  \caption{\textbf{Shape evolution on Waymo.} In (a), from left to right is the results of shape reconstruction at different tracking steps. The ground-truth aggregated point clouds are shown in (b). The shapes are become more and more aligned with ground-truth point clouds via online adaptation. }
  \label{fig: evolution}
  \vspace{-0.1in}
\end{figure}

\subsection{Comprehensive Comparisons}

\noindent
\textbf{Evaluations on Waymo.} We compare our method with SOTracker, a recent single 3D object tracking method. As shown in Table~\ref{tab: comparison_waymo}, our method achieves comparable results on the easy subset and outperforms SOTracker by a large margin on the hard subset. It is reasonable that shape reconstruction is more beneficial in the case of more sparse point clouds, since the shape provides a global description of the object which is not prone to the error of partial and incomplete view.

We also evaluate the performance of the shape reconstruction. Since shapes are represented by aggregated point clouds in SOTracker, metrics including ACD and Recall are not suitable, we first aggregate point clouds based on the predicted pose and then exploit our SDF model to convert the point clouds to mesh for comparison. As shown in Table~\ref{tab: comparison_waymo}, our method outperforms the baseline on every subset. This experiment demonstrates that joint tracking and shape reconstruction leads to better performance than tracking followed by shape reconstruction.

\noindent
\textbf{Evaluations on KITTI.} We further evaluate our method on the KITTI dataset. We compare our method with SC3D\cite{giancola2019leveraging}, P2B\cite{qi2020p2b}, 3D-SiamRPN\cite{Fang20213DSiamRPNAE}, MLVSNet\cite{wang2021mlvsnet}, BAT\cite{zheng2021box} and SOTracker\cite{pang2021model}, which are recent single object tracking methods on the KITTI dataset.  Since all learning-based methods employ the annotated bounding box for training while our method only pretrains on ShapeNet, we employ the detection term defined in Eq.~\ref{eqn: detection} on the KITTI dataset. Specifically, we utilize the predicted bounding box provided by VoxelRCNN~\cite{DBLP:conf/aaai/DengSLZZL21}. For a fair comparison, we also enable the detection term of SOTracker.
In addition, we introduce a baseline using the Kalman filter to associate detection bounding boxes (VoxelRCNN + KF in Table~\ref{tab: comparison_kitti}).

As shown in Table~\ref{tab: comparison_kitti}, our method outperforms most of previous methods and achieves comparable performance with the state-of-the-art method BAT, and outperforms SOTracker and VoxelRCNN + KF by a large margin. Similarly, we aggregate point clouds and convert them to mesh to evaluate the performance of the shape reconstruction. We again demonstrate the advantages of joint tracking and shape reconstruction. Note that the protocol of the KITTI dataset differs from LiDAR-SOT. In the beginning frames of some tracklets, the point clouds are highly sparse, which is the reason why the optimization-based methods (SOTracker and Ours) do not achieve the expected performance.

\begin{table*}[t]
  \centering
  \begin{tabular}{lcccccccc}
    \toprule
    & \multicolumn{2}{c}{All} & \multicolumn{2}{c}{ Easy } & \multicolumn{2}{c}{Medium} & \multicolumn{2}{c}{Hard} \\
     & Succ $\uparrow$ & Prec $\uparrow$ & Succ $\uparrow$ & Prec $\uparrow$ & Succ $\uparrow$ & Prec $\uparrow$ & Succ $\uparrow$ & Prec $\uparrow$ \\
    \midrule
    w/o regularizer & 58.1 & 61.3 & 67.8 & 70.0 & 53.6 & 56.5 & 51.3 & 56.2 \\
    w/o CD loss & 60.0 & 63.4 & 68.3 & 70.4 & 56.5 & 59.3 & 54.0 & 59.3 \\
    w/o shape loss & 42.2 & 42.4 & 57.5 & 57.8 & 34.2 & 34.0 & 32.4 & 33.1 \\
    full model & \textbf{62.3} & \textbf{65.7} & \textbf{71.5} & \textbf{74.1} & \textbf{58.8} & \textbf{61.8} & \textbf{54.9} & \textbf{59.9} \\
    \bottomrule
  \end{tabular}
  \caption{\textbf{Ablation study on Waymo.} We analyze the regularizer, the Chamfer distance loss and the shape loss.}
  \label{tab: ablation_waymo}
\end{table*}

\begin{figure*}[t]
  \centering
  \vspace{-0.1in}
  \includegraphics[width=0.9\textwidth]{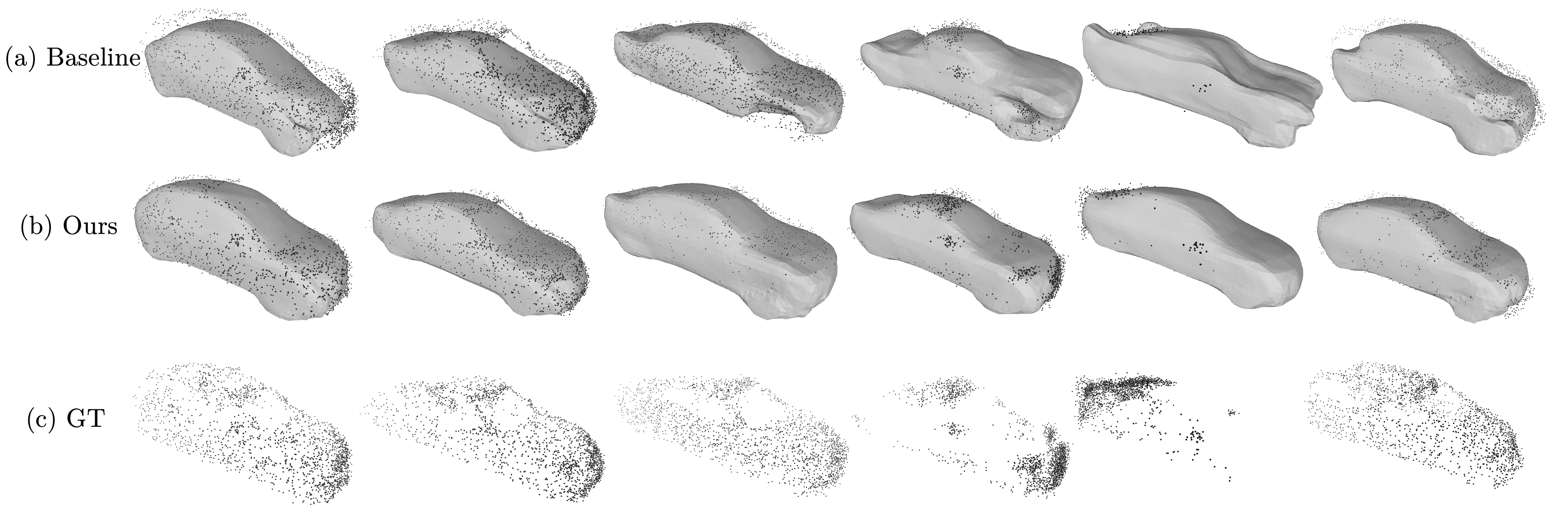}
  \vspace{-0.05in}
  \caption{\textbf{Reconstructed shape from Waymo.} Our method produces more accurate shape than the baseline. The ground-truth aggregated point clouds are shown in (c).}
  \label{fig: reconstruction}
  \vspace{-0.15in}
\end{figure*}

\begin{table}[t]
  \centering
  \begin{tabular}{lcc}
    \toprule
     & Succ $\uparrow$ & Prec $\uparrow$ \\
    \midrule
    w/o detection loss & 42.3 & 52.4 \\
    w/o shape loss & 58.7 & 73.7 \\
    full model & \textbf{62.3} & \textbf{77.7} \\
    \bottomrule
  \end{tabular}
  \caption{\textbf{Ablation study on KITTI.} We analyze the contributions of detection loss and shape loss.}
  \label{tab: ablation_kitti}
  \vspace{-0.1in}
\end{table}

\begin{figure}[t]
  \centering
  \includegraphics[width=1.0\columnwidth]{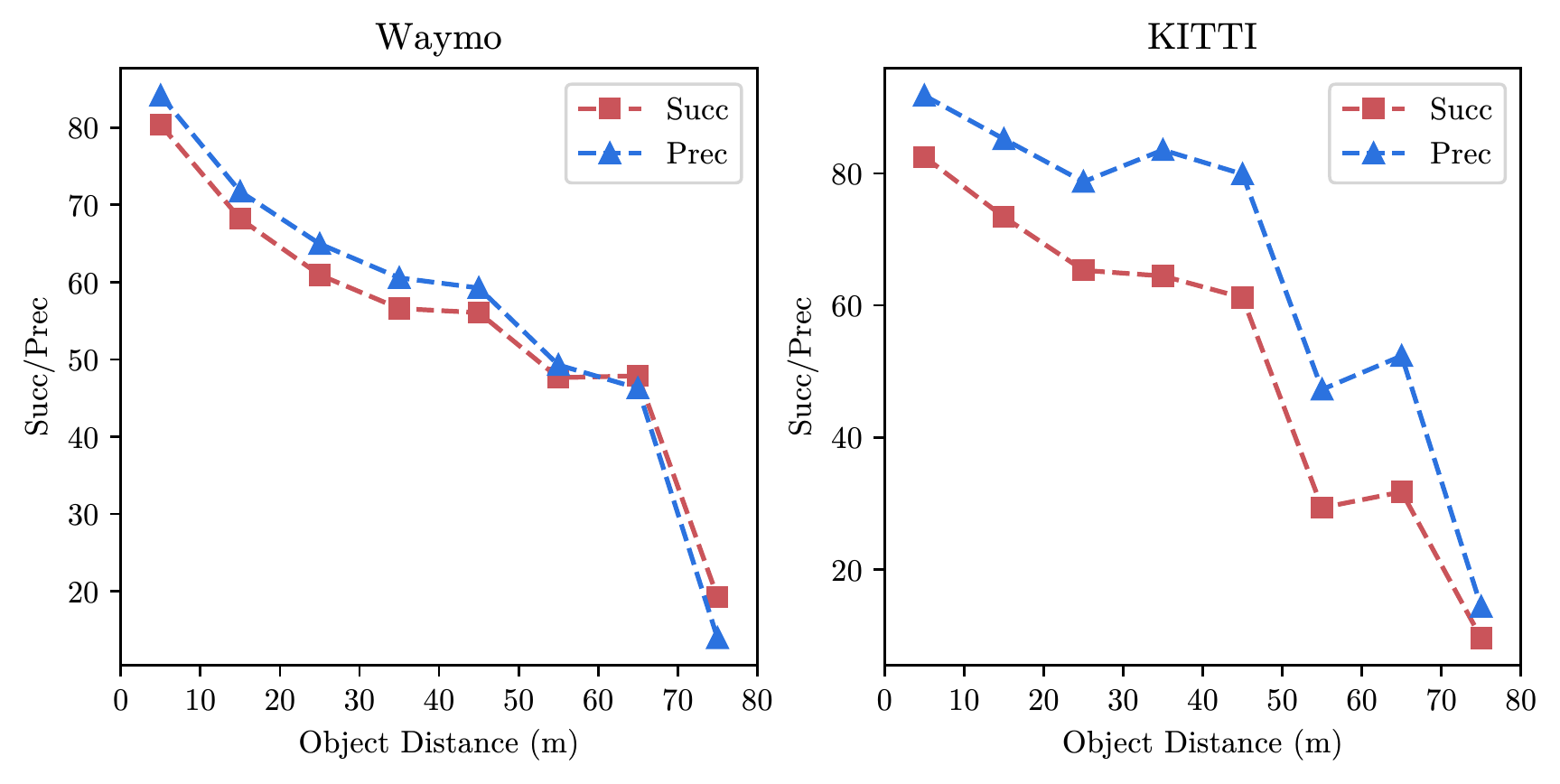}
  \vspace{-0.2in}
  \caption{\textbf{Analysis of object distances.} As the distance increases, the performance decreases. However, our method still maintains a reasonable performance within 60 meters. }
  \label{fig: distance}
  \vspace{-0.15in}
\end{figure}

\subsection{Ablation Study}

\noindent
\textbf{Effectiveness of online adaptation mechanisms.} To demonstrate the effectiveness of the online adaptation mechanism, we only adapt the shape in the first  several frames of different lengths starting at each tracklet and fix the shape code in all subsequent frames. Then we compare the performance of tracking and reconstruction. We conduct this experiment on the Waymo dataset and exploit only shape loss for tracking and reconstruction.
Besides, we extract the mean shape from ShapeNet and perform tracking with it (denoted as updating 0 frames in the figure).

As shown in Figure~\ref{fig: adaptation_tracking}, the tracking metric on the Waymo dataset continues to improve as the number of adapted frames increases. Similarly, we evaluate the performance of shape reconstruction in the case of different numbers of update frames. As shown in Figure~\ref{fig: adaptation_shape}, online adaptation also improves the quality of the shape. Figure~\ref{fig: evolution} illustrates our shape evolution during the tracking process on the Waymo dataset. We also find that as the number of updates increases, the gains for tracking and reconstruction become saturated.

\noindent
\textbf{Ablation study.} We conduct thorough ablation studies on the Waymo and the KITTI datasets. We report our results in Table~\ref{tab: ablation_waymo} and Table~\ref{tab: ablation_kitti}. We ablate the regularizer ($\ell_2$ regularizer in Eq.~\ref{eqn: adaptation}),
the shape loss (first term in Eq.~\ref{eqn: tracking} and Eq.~\ref{eqn: adaptation}) and the Chamfer distance loss (CD loss, second term in Eq.~\ref{eqn: tracking}) and observe all of them can improve the tracking performance. Among them, the proposed shape loss plays a far more important role than the Chamfer distance loss. We also observe applying object detection significantly boosts performance in the KITTI dataset given the noisy and sparse point cloud inputs.

\noindent
\textbf{Analysis of object distance.} We measure the tracking performance of objects at different distances. Specifically, we calculate the average distance of the object during the whole tracking process and divide them into different bins for every 10 meters. We show the result in Figure~\ref{fig: distance}. From the figure, we can tell that as the distance increases, the performance decreases. However, our method still maintains a reasonable performance within 60 meters.

\subsection{Qualitative Analysis}

Figure~\ref{fig: reconstruction} compares our reconstruction results with the baseline on the Waymo dataset. For the baseline, we use the LiDAR-SOT~\cite{pang2021model} to aggregate the point clouds over time and apply the DeepSDF on the aggregated points to reconstruct shape at once. Figure~\ref{fig: reconstruction} shows that our method produces more accurate shapes compared to the baseline. Due to the shape prior and the online adaptation mechanism, even when the input is sparse and partial, our method is still able to produce an acceptable shape.

\section{Conclusion and Limitation}

\textbf{Conclusion.} In this paper, we present a novel and unified framework for object tracking and shape reconstruction in the wild. We propose to leverage the continuity and redundancy in video data with a differentiable shape model. Specifically, we utilize a DeepSDF model to simultaneously perform object tracking and 3D reconstruction. During the tracking process, we adapt the shape model based on new observations to improve the shape quality, which leads to improvement on tracking and vice versa. We also demonstrate that the results on both Waymo and KITTI datasets outperform state-of-the-art methods by a large margin.

\textbf{Limitation.} We describe two limitations of our method: (i) Since we have to focus on both object tracking and shape reconstruction, shapes can not be overfitted to the observation at the current time step. Otherwise, it would be difficult to generalize to subsequent novel views. Hence, we have to maintain a relatively smooth shape during online tracking. (ii) Although we propose the implicit shape to reduce the accumulated error, the drifting issue still exists and future research is required.

\ifCLASSOPTIONcaptionsoff
  \newpage
\fi

\bibliography{IEEEabrv,reference_simplified.bib}
\bibliographystyle{IEEEtran}

\addtolength{\textheight}{-12cm}   %
\end{document}